\documentclass[acmtog,authorversion]{acmart}

\usepackage{amsmath}

\DeclareMathOperator*{\argmin}{argmin}
\usepackage{bm}
\usepackage{cleveref}
\usepackage{multirow}
\usepackage{float}
\usepackage{bbding}

\newcommand{\revised}[1]{{#1}}

\setcopyright{acmlicensed}
\acmJournal{TOG}
\acmYear{2024} \acmVolume{43} \acmNumber{6} \acmArticle{} \acmMonth{12}\acmDOI{10.1145/3687961}

\begin{document}

\title{Accelerate Neural Subspace-Based Reduced-Order Solver of Deformable Simulation by Lipschitz Optimization}

\author{Aoran Lyu}
\authornote{Equal contribution of the first two authors.}
\email{lvaoran@hotmail.com}
\orcid{0000-0002-5142-5979}
\affiliation{
  \institution{South China University of Technology}
  \country{China / The University of Manchester, United Kingdom}
}

\author{Shixian Zhao}
\authornotemark[1]
\email{cssxzhao@mail.scut.edu.cn}
\orcid{0009-0004-8677-2331}
\affiliation{
  \institution{South China University of Technology}
  \city{Guangzhou}
  \country{China}
}

\author{Chuhua Xian}
\authornote{Corresponding authors.}
\email{chhxian@scut.edu.cn}
\orcid{0000-0001-7656-4652}
\affiliation{
  \institution{South China University of Technology}
  \city{Guangzhou}
  \country{China}
}

\author{Zhihao Cen}
\email{czh1224415633@gmail.com}
\orcid{0009-0009-5967-3895}
\affiliation{
  \institution{South China University of Technology}
  \city{Guangzhou}
  \country{China}
}

\author{Hongmin Cai}
\email{hmcai@scut.edu.cn}
\orcid{0000-0002-2747-7234}
\affiliation{
  \institution{South China University of Technology}
  \city{Guangzhou}
  \country{China}
}

\author{Guoxin Fang}
\authornotemark[2]
\email{guoxinfang@cuhk.edu.hk}
\orcid{0000-0001-8741-3227}
\affiliation{
  \institution{The Chinese University of Hong Kong}
  \city{Hong Kong SAR}
  \country{China}
}

\begin{abstract}
Reduced-order simulation is an emerging method for accelerating physical simulations with high DOFs, and recently developed neural-network-based methods with nonlinear subspaces have been proven effective in diverse applications as more concise subspaces can be detected. However, the complexity and landscape of simulation objectives within the subspace have not been optimized, which leaves room for enhancement of the convergence speed. This work focuses on this point by proposing a general method for finding optimized subspace mappings, enabling further acceleration of neural reduced-order simulations while capturing comprehensive representations of the configuration manifolds. We achieve this by optimizing the Lipschitz energy of the elasticity term in the simulation objective, and incorporating the cubature approximation into the training process to manage the high memory and time demands associated with optimizing the newly introduced energy. Our method is versatile and applicable to both supervised and unsupervised settings for optimizing the parameterizations of the configuration manifolds. We demonstrate the effectiveness of our approach through general cases in both quasi-static and dynamics simulations. Our method achieves acceleration factors of up to 6.83 while consistently preserving comparable simulation accuracy in various cases, including large twisting, bending, and rotational deformations with collision handling. This novel approach offers significant potential for accelerating physical simulations, and can be a good add-on to existing neural-network-based solutions in modeling complex deformable objects. 
\end{abstract}




\begin{teaserfigure}
  \centering
  \includegraphics[width=0.99\linewidth]{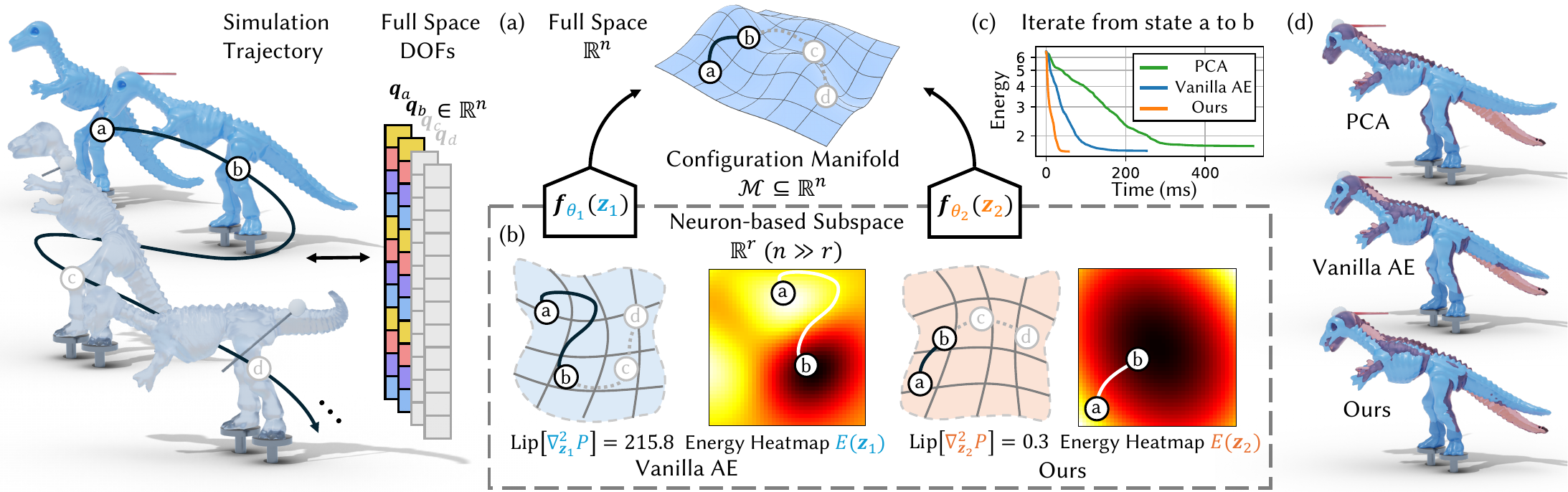}
      \caption{We propose a Lipschitz optimization method that can significantly accelerate the convergence speed of reduced-order simulations driven by neural-network-based approaches. (a) The deformation process can be formulated as a path through a configuration manifold $\mathcal{M} \subseteq \mathbb{R}^n$, where reduced-order solvers tend to find a mapping $\bm{f}_\theta(\bm{z})$ that maps a low-dimensional subspace $\mathbb{R}^r$ to the manifold. (b) Our method enhances the objective landscape in the neural subspace by minimizing the second-order Lipschitz regularization energy, which substantially improves convergence speed when using iterative solvers like Newton's method. (c, d) Compared to conventional linear subspace methods (driven by PCA) and direct neural subspace constructions, our method achieves faster convergence and maintains quality when using the same subspace dimension.}
  \label{fig-teaser}
\end{teaserfigure}

\maketitle

\section{Introduction}

Physical modeling of deformable objects is a significant aspect of computer graphics developments. When aiming at achieving high-detail and realistic simulation, the system generally contains large degrees of freedom (DOFs), and therefore applying numerical solvers (e.g., the implicit Euler with Netwon's or L-BFGS method) in the \textit{full space} $\mathbb{R}^n$ to capture statics/dynamics properties by minimizing nonlinear simulation objectives can be time-consuming. Reduced-order solvers provide a promising solution to lower the computational time while maintaining the accuracy, and the key is to find a $r$-dimensional ($r \ll n$) configuration manifold $\mathcal{M} \subseteq \mathbb{R}^n$ that can capture the essential features of the deformation patterns in $\mathbb{R}^n$. As illustrated in Fig.~\ref{fig-teaser}(b), a subspace mapping function $\bm{f}_\theta:\mathbb{R}^r\rightarrow\mathbb{R}^n$ needs to be identified correspondingly. In the reduced-order solver, the iterative-based numerical process operates within the lower-dimensional subspace, therefore greatly reducing the computational cost and iteration time. 

Finding a concise subspace and an efficient mapping function 
are two critical aspects for optimizing the performance (computational time and accuracy) of reduced-order solvers; however, balancing these two remains challenging. Conventional methods adopt linear mapping functions, and a widely applied method is to detect subspaces by considering the principal components of perturbative motions to extract the most significant modes of deformation variation ~\cite{pentland1989good}. With the usage of \textit{linear} $\bm{f}_\theta$, the solver can be computationally efficient as a simple evaluation of the subspace Jacobian/Hessian of the objective can be obtained. While this can also reduce the iteration time, these methods may need a large subspace to capture complex nonlinear deformations in the full space ~\cite{benchekroun2023fast}, leading to recent developments in nonlinear \revised{neural} subspace mapping. With the aid of \textit{multi-layer perceptions} (MLP) and the differentiable optimization pipeline with network back-propagation, a more concise subspace can be detected and therefore further lower the dimension of computation within the subspace~\cite{fulton2019latent}. However, existing methods generally overlook the drawback that comes from the complexity of the objective function in subspace - without a carefully selected subspace mapping, the subspace Hessian matrices cannot well capture the local information therefore will bring slow convergence on iterative solvers like Newton's method~\cite{sharp2023data}. 
It still lacks of a general study on the objective landscape of the subspace to further improve the performance of \revised{neural} reduced solvers in physical simulation. 

\subsection{Our Method}

In this work, we focus on identifying a nonlinear \revised{neural} subspace that is optimized to reduce the complexity of the objective within the subspace (i.e. improving the objective landscape),
thereby leading to faster convergence in the reduced-order solver. We have observed that the Lipschitz characteristics of the subspace-objective mapping significantly influence performance (as illustrated in Fig.~\ref{fig-teaser}(b)). Additionally, the flexibility of network mapping provides ample space for optimization, allowing for the identification of effective mapping functions that achieve similar or identical outcomes as the original, but with improved Lipschitz characteristics for the objective. We propose performing Lipschitz optimization of the simulation objective during the construction of neural subspace mappings. Thanks to recent advancements in neural subspace mapping ~\cite{fulton2019latent, shen2021high, sharp2023data}, we can integrate our Lipschitz optimization method into existing neural subspace construction schemes to achieve further speedups without compromising result quality. The Lipschitz optimization is carried out by varying the coordinate form of the metric through different parameterizations of the manifold. Our approach involves adding a self-supervised loss term to the training process and implementing an optional cubature acceleration. At runtime, the acceleration of the neural reduced-order simulation can be realized without the need for additional network structures or new computational processes.

The technical contributions are summarized as follows:
\begin{itemize}

    \item We propose a general method to optimize the objective landscape in nonlinear \revised{neural} subspaces, which can accelerate the convergence of reduced-order solvers for deformable objects. Our method is applicable to both supervised and unsupervised cases.

    \item We incorporate cubature approximation to significantly reduce memory and time costs when optimizing the 2nd-order Lipschitz energy, ensuring successful learning for cases with large DOFs.
    
\end{itemize}

We demonstrate the effectiveness of our approach in different simulation cases. By performing Lipschitz optimization while preserving simulation accuracy, our method achieves an acceleration factor ranging from 1.42 to 6.83 across various cases, including large twisting, bending, and rotational deformations with collision handling. 

\section{Related Work}

We review the literature on reduced-order methods for efficiently conducting physical simulations. Additionally, as the key observation of this work, we explore related work on Lipschitz regularization for neural networks and their graphics applications.

\subsubsection*{Deformable Simulation by Numerical Solver}
Notable achievements in the graphics community have been developed, which support simulations of elastic objects with large deformations ~\cite{terzopoulos1987elastically, irving2004invertible}, consider nonlinear material properties~\cite{baraff2023large, bonet1997nonlinear, smith2018stable} and can handle fast and accurate collision response~\cite{li2020incremental, wang2023fast}.
Implicit Euler method with Newton's solver has been proven to have good convergence for time integration of the simulation system. However, optimizing the non-convex elastic energy can result in high computational costs, which limits the feasibility of conducting real-time simulations. To accelerate these processes, a large number of methods have been proposed, including multigrid solvers (e.g., ~\cite{tamstorf2015smoothed}), fast Hessian projections (e.g., ~\cite{mcadams2011efficient, smith2019analytic}), quasi-Newton methods that avoid direct evaluation of Hessians (e.g., ~\cite{peng2018anderson}), and project dynamics (e.g., ~\cite{bouaziz2014projective}) which can be seen as a type of quasi-Newton methods ~\cite{liu2017quasi}. In this work, we aim to improve the convergence speed of Newton's method by optimizing the landscape of the  \revised{neural} reduced solver.

\subsubsection*{Reduce-order Methods for Simulation}
As another appealing acceleration solution, the reduced order method achieve speed up by constraining the possible states to a low-dimensional configuration manifold immersed in the high-dimensional \textit{full space}~\cite{hahn2014subspace, kim2011physics}. 
\revised{In the early stage of development, linear mapping functions detected by modal analysis}~\cite{pentland1989good, james2002dyrt, Hauser2003InteractiveDU} \revised{through the selection of features based on the elastic potential Hessian is used.}
To better mimic large rotational and twisting deformations, subspace mappings based on principal component analysis (PCA)~\cite{krysl2001dimensional} or linear blend skinning~\cite{von2013efficient, brandt2018hyper, benchekroun2023fast, trusty2023subspace} have been introduced. However, methods based on linear mapping generally require a large subspace dimension to handle complex and nonlinear deformations effectively.

Recently, \revised{neural} nonlinear subspace mappings showed their superiority in representing rich deformations with a compact subspace ~\cite{fulton2019latent, zong2023neural, sharp2023data}. 
However, the exceeding non-linearity makes these mappings suffer from over-fitting the configuration manifolds~\cite{shen2021high} and high iteration numbers~\cite{sharp2023data}. The problems caused by exceeding non-linearity were partially solved by combining the neural mappings with a linear mapping that runs sequential~\cite{fulton2019latent} or parallel~\cite{shen2021high} to the neural part.
Meanwhile, to achieve further acceleration of general reduced-order simulations, cubature methods ~\cite{an2008optimizing, von2013efficient, yang2015expediting, trusty2023subspace} were proposed to approximate reduced elastic force using only a subset of cubature samples on the deformable. 
Our work present a general study on the objective landscape of neural subspace mappings to improve the simulation speed, and the cubature method is used during the mapping construction to drastically reduce the cost of 2nd-order Lipschitz regularization. 

\subsubsection*{Lipschitz Regularization in Subspace construction}
The input-output smoothness of neural networks (e.g., the landscape of subspace mapping characterized by zero-order Lipschitz energy) has been recognized to correlate with the generalization ability and robustness of the network ~\cite{Hoffman2019RobustLW, simon2019first}. To encourage the input-output smoothness, one can penalize the norm of the Jacobian of the input-output mapping ~\cite{jakubovitz2018improving, gulrajani2017improved, terjek2019adversarial}. There also exist methods precisely bound the network's Lipschitz constant by modifying the weight matrices ~\cite{miyato2018spectral, gouk2021regularisation, anil2019sorting}. \revised{This type of technique was also introduced to graphics} ~\cite{liu2022learning} to improve the smoothness of neural-field-based shape representations with an improvement on automatically learning an appropriate Lipschitz bound~\cite{qin2019adversarial, moosavi2019robustness}. 
\revised{In the context of reduced-order simulations, Chen et al.~\shortcite{chen2023model} proposed a technique to enhance the accuracy of projection-based methods by regularizing the subspace mapping. They introduced a velocity loss term that penalizes errors from linear approximations derived from the Jacobian of the subspace mapping across adjacent timesteps. Similarly, our work also implements regularization of a mapping. In contrast, since our goal is to accelerate simulation speed, we apply regularization to the input-objective Hessian mapping by minimizing the second-order Lipschitz constant.}
(detail discussed in Sec.~\ref{sec-convergence-discussion} and Sec.~\ref{sec-method}).

\section{\revised{Neural} Reduced Order Solver: Preliminary and Analysis}

\begin{figure}[t]
  \centering
  \includegraphics[width=1\linewidth]{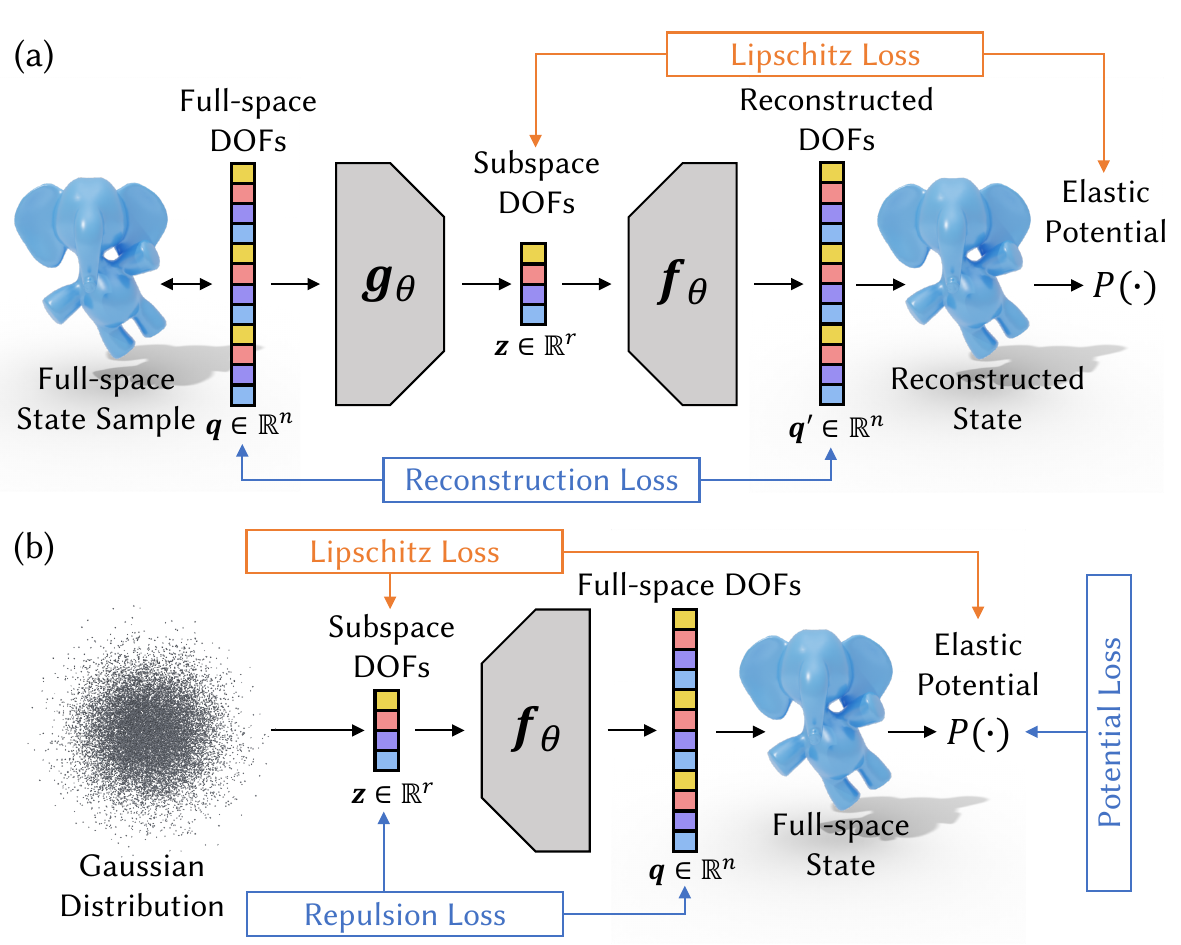}
  \caption{Network training settings for effective  \revised{neural} subspace construction. (a) The supervised setting. (b) The unsupervised setting. Conventional methods only consider the loss shown in blue but do not optimize the Lipschitz loss (shown in orange) to control the landscape of simulation objective in the subspace.}
  \label{fig-training-settings}
\end{figure}

In this section, we give the background information of reduced order solvers and conventional training processes for detecting \revised{neural} subspace and its mapping function, followed by a short discussion on the issue of convergence speed of existing methods.

\subsection{Reduced-order Solvers}
Given the discretization of the deformables, the system configuration \revised{in deformable simulations} can be represented by a DOF vector $\bm{q} \in \mathbb{R}^n$ (e.g., vertex positions), which also gives a parameterization of the $n$-dimensional full space. \revised{The system configuration defines the current deformation of the deformable object and the elastic energy density induces the mapping from strain (deformation) to stress. For simulations considering kinetic properties, the time evolution of the system can be obtained by time integrating the governing equations. In the need for stability, the implicit Euler integration is often employed as the integration scheme}, which can be applied by solving an optimization problem ~\cite{martin2011example, bouaziz2014projective, liu2017quasi}:
\begin{equation}
\label{form-full-space-time-integration}
    \bm{q}^{k+1}=\argmin_{\bm{q}}\left[\frac{1}{2\Delta t^2}||\bm{q}-\overline{\bm{q}}^{k+1}||^2_{\bm{M}} + P(\bm{q})\right],
\end{equation}
where $\bm{q}^{k+1}$ is the DOF vector of timestep $k+1$, $\Delta t$ is the timestep, $\overline{\bm{q}}^{k+1}$ is the inertial guess determined by previous configurations and the external force, $\bm{M}$ is the lumped mass matrix, and $P(\cdot)$ is the elastic potential\revised{, which can be obtained by integrating the elastic energy density. We refer to \cite{kim2022dynamic} for a comprehensive introduction to deformable simulations.} 

When applying reduced-order method, a subspace mapping is \revised{introduced} as $\bm{f}(\bm{z}): \Omega \rightarrow \mathcal{M}$, which maps the low-dimensional subspace coordinates $\bm{z} \in \Omega \subseteq \mathbb{R}^r$ to a $r$-dimensional configuration manifold $\mathcal{M}={\rm{Im}}(\bm{f})$ immersed in the high-dimensional full space $\mathbb{R}^n$.
The optimization problem can then be reformulated as:
\begin{equation}
\label{form-reduced-space-time-integration}
    \bm{z}^{k+1}=\argmin_{\bm{z}} E(\bm{z}) = \argmin_{\bm{z}}\left[\frac{1}{2\Delta t^2}||\bm{f}(\bm{z})-\overline{\bm{q}}^{k+1}||^2_{\bm{M}} + P(\bm{f}(\bm{z}))\right],
\end{equation}
where $\bm{z}^{k+1}$ is the subspace coordinates $\bm{z}$ of timestep $k+1$. The optimization problem is generally solved using iterative solvers such as Newton's and quasi-Netwon methods (if approximated Hessians are adopted). It is worth mentioning that the elasticity term provides the major nonlinearity of the optimization problem, and when performing quasi-static simulations, the inertia term in $E(z)$ is omitted. 

\subsection{\revised{Learning Neural Subspaces}}

For the  \revised{neural} representation of subspace, the mappings' function space is parameterized by the network weights (denoted as $\theta$). Existing construction methods can be categorized into two classes: supervised setting and unsupervised setting.
\subsubsection*{Supervised Setting}
When a set of observations of the system configurations $\mathcal{Q}=\{\bm{q}_i \in \mathbb{R}^n\}$ (i.e. a dataset) is available, one can build and train an autoencoder structure consists of an encoder $\bm{g}_{\theta}: \mathbb{R}^n \rightarrow \mathbb{R}^r$ and a decoder $\bm{f}_{\theta}: \mathbb{R}^r \rightarrow \mathbb{R}^n$. The decoder will be the subspace mapping for the reduced-order simulation. To make the decoder's image ${\rm{Im}}(\bm{f}_{\theta})$ cover the configuration observations, the parameter $\theta$ is obtained by training the network with the \textit{reconstruction loss} ~\cite{fulton2019latent, shen2021high} (see Fig.~\ref{fig-training-settings}(a)):
\begin{equation}
\label{form-vanilla-supervised-construction}
\min_\theta \frac{1}{|\mathcal{Q}|}\sum_{\bm{q}_i \in \mathcal{Q}}||\bm{f}_\theta(\bm{g}_\theta(\bm{q}_i))-\bm{q}_i||_{\bm{M}}^2.
\end{equation}
\subsubsection*{Unsupervised Setting}
Recently, Sharp \textit{el al.}~\shortcite{sharp2023data} proposed an unsupervised method to construct a neural subspace without the need for a dataset. The method trains the mapping $\bm{f}_{\theta}$ from an energy-first perspective like the Linear Model Analysis ~\cite{pentland1989good}:
\begin{equation}
\label{form-vanilla-unsupervised-construction}
\min_\theta \left[ \mathbb{E}_{\bm{z}\sim \mathcal{N}}[P(\bm{f}_\theta(\bm{z}))] + \lambda \mathbb{E}_{\bm{z}_1, \bm{z}_2 \sim \mathcal{N}}\left[ \log \left(\frac{||\bm{f}_\theta(\bm{z}_1)-\bm{f}_\theta(\bm{z}_2)||_{\bm{M}}}{\sigma||\bm{z}_1 - \bm{z}_2||}\right)^2 \right] \right],
\end{equation}
where $\mathcal{N}$ is the standard Gaussian distribution, $\lambda$ and $\sigma$ are 2 hyper-parameters. Here, the first term is a \textit{potential loss} that encourages the image ${\rm{Im}}(\bm{f}_{\theta})$ concentrate on low-energy profiles and the second term is a \textit{repulsion loss} that prevents the image from collapsing to a point (see Fig.~\ref{fig-training-settings}(b)).

For both settings, the subspace mapping $\bm{f}$ generally comes from a parameterized function space \revised{$\mathcal{F} =\{\bm{f}_\theta: \forall \theta\}$}. Once the parameter $\theta$ as the network weights are determined, the mapping $\bm{f}_{\theta}$ is also constructed. For ease of expression, we \revised{denote} all loss functions in \cref{form-vanilla-supervised-construction} and \cref{form-vanilla-unsupervised-construction} as $\mathcal{L}_{C}(\theta)$.

\subsection{Short Discussion on Convergence Speed}
\label{sec-convergence-discussion}

The cost of solving optimization problem \cref{form-reduced-space-time-integration} with nonlinear  \revised{neural} subspace using iterative methods like Newton's method is governed by three factors: cost of evaluating objective and its derivatives $C_{\rm{eval}}$, cost of finding appropriate stepping directions $C_{\rm{dir}}$ (by using Hessian or its estimations), and number of stepping iterations $n_{\rm{iter}}$. The overall cost can be roughly estimated as $n_{\rm{iter}}(C_{\rm{eval}} + C_{\rm{dir}})$. With the help of a  \revised{neural} reduced solver, the $C_{\rm{dir}}$ can be greatly reduced due to the low problem dimension $r \ll n$. On the other hand, the convergence of the solver (i.e., the number of iterations) highly depends on the properties of the problem. In particular when optimizing $E(z)$ with Lipschitz continuous Hessian $\nabla_{\bm{z}}^2 E$ using Newton's method, we have a quadratic convergence rate~\cite{nocedal2006numerical} written as:
\begin{equation}
\|\bm{z}^{k+1} - \bm{z}^*\| \leq {\rm{Lip}}[\nabla_{\bm{z}}^2 E] \left\| \nabla_{\bm{z}}^2 E(\bm{z}^*)^{-1} \right\| \|\bm{z}^{k} - \bm{z}^*\|^2,
\end{equation}
where $\bm{z}^{k+1}$ and $\bm{z}^{k}$ are the $(k+1)$-th and the $k$-th Newton iteration result, respectively, $\bm{z}^*=\argmin E(\bm{z})$ is a solution, and ${\rm{Lip}}[\nabla_{\bm{z}}^2 E]$ is the Lipschitz constant of the problem Hessian $\nabla_{\bm{z}}^2 E$. It can be seen that, the number of iterations required to reach a certain error threshold scales with the Hessian's Lipschitz constant ${\rm{Lip}}[\nabla_{\bm{z}}^2 E]$:
\begin{equation}
\label{form-Hessian-Lipschitz}
{\rm{Lip}}[\nabla_{\bm{z}}^2 E] = \max_{\bm{z}_1, \bm{z}_2 \in \Omega} \frac{||\nabla_{\bm{z}}^2 E(\bm{z}_1) - \nabla_{\bm{z}}^2 E(\bm{z}_2)||}{||\bm{z}_1 - \bm{z}_2||},
\end{equation}
where $\Omega$ is the domain of the subspace mapping $\bm{f}$. In this work, we focus on accelerating the simulation by reducing the number of iterations $n_{\rm{iter}}$ through the use of our Lipschitz regularization in subspace construction. Our key observation is that, with a given subspace mapping $\bm{f}_{\theta^{\rm{init}}} \in \mathcal{F}$ \revised{obtained from minimizing the construction loss $\mathcal{L}_C(\theta^{\rm{init}})$}
, it is possible to optimize ${\rm{Lip}}[\nabla_{\bm{z}}^2 E]$ by finding another parameter $\theta^*$ with smaller ${\rm{Lip}}[\nabla_{\bm{z}}^2 E]$ while preserving the image of the mapping \revised{(i.e., making ${\rm{Im}}(\bm{f}_{\theta^*})\approx{\rm{Im}}(\bm{f}_{\theta^{\rm{init}}})$)}. We tend to achieve this by define and reduce a loss approximating the second-order Lipschitz constraint ${\rm{Lip}}[\nabla_{\bm{z}}^2 E]$ together with $\mathcal{L}_C$.
\revised{As $\mathcal{L}_C(\theta^*)$ characterizes the image ${\rm{Im}}(\bm{f}_{\theta^*})$, we retain the similarity of ${\rm{Im}}(\bm{f}_{\theta^*})$ and ${\rm{Im}}(\bm{f}_{\theta^{\rm{init}}})$ by keeping $\mathcal{L}_C(\theta^*)$ in the optimization objective of $\theta^*$, which is verified in the numerical test (presented in Sec.~\ref{section-supervised-dynamics-results})}.
The loss function and training details with cubature acceleration are presented in the next section.

\section{Loss with Lipschitz Optimization and Training Details}
\label{sec-method}

We now discuss the details of deriving our Lipschitz loss (\revised{denoted} as $\mathcal{L}_{LS}$) for training, and details on estimating the Hessian to accelerate the training process and reduce the memory usage.

\subsection{Lipschitz Loss}

Directly optimizing the Lipschitz constant of problem Hessian \cref{form-Hessian-Lipschitz} is intractable since exact computation requires traversing all possible point pairs in the domain. To make it computable and optimizable, one can approximate it using the order statistics from a set of observations $\mathcal{Z}=\{\bm{z}_i \overset{\mathrm{iid}}{\sim} \revised{\Pi_\theta}(\bm{z})\}$ where $\revised{\Pi_\theta}(\bm{z})$ is the pull-back distribution induced from the configuration distribution:
\begin{equation}
\label{form-H-Lipschitz-estimation}
{\rm{Lip}}[\nabla_{\bm{z}}^2 E]\approx\max_{\bm{z}_i, \bm{z}_j \in \mathcal{Z}, \bm{z}_i \neq \bm{z}_j}\left\{ \frac{||\nabla_{\bm{z}}^2 E(\bm{z}_i) - \nabla_{\bm{z}}^2 E(\bm{z}_j)||}{||\bm{z}_i - \bm{z}_j||} \right\}.
\end{equation}
However, the gradients produced by this approximation are spatially very sparse, specifically, each Lipschitz optimizing iteration will only optimize against two points in the subspace and may damage the Lipschitz characteristics around other locations. Noticing that the $\max$ operator can be softened to a $p$-norm:
\begin{equation}
{\rm{Lip}}[\nabla_{\bm{z}}^2 E]\approx\left(\sum_{\bm{z}_i, \bm{z}_j \in \mathcal{Z}, \bm{z}_i \neq \bm{z}_j} \frac{||\nabla_{\bm{z}}^2 E(\bm{z}_i) - \nabla_{\bm{z}}^2 E(\bm{z}_j)||^p}{||\bm{z}_i - \bm{z}_j||^p} \right)^{\frac{1}{p}}.
\end{equation}
Here, the $\max$ operator can be recovered when $p \rightarrow \infty$. By taking $p=2$, omitting the sample-independent exponent $1/p$, and adding a multiplier $1/(n(n-1))$ making the term independent to the sample number, we obtain a \textit{Lipschitz loss} related to the Hessian's Lipschitz constant:
\begin{equation}
\label{form-L-LS-E}
\mathcal{L}_{\rm{LS}} = \mathbb{E}_{\bm{z}_1, \bm{z}_2 \overset{\mathrm{iid}}{\sim} \revised{\Pi_\theta}(\bm{z})}\left[ 
\frac{||\nabla_{\bm{z}}^2 E(\bm{z}_1) - \nabla_{\bm{z}}^2 E(\bm{z}_2)||^2}{||\bm{z}_1 - \bm{z}_2||^2} \right],
\end{equation}
which can be calculated over a batch of subspace samples and provide spatially dense gradients for optimization. \revised{Note that this loss can be generalized for Lipschitz constants of arbitrary orders, and we refer to the supplementary document for more discussions. In this work, we consider systems with Lipschitz continuous subspace Hessians (i.e., deformation processes involving plasticity and fractures are excluded). }

\subsection{Training Details}
\label{sec-training-details}

To perform our Lipschitz optimization of problem \cref{form-reduced-space-time-integration}, a parameterized function space $\mathcal{F}$ of subspace mappings as well as a mapping $\bm{f}_{\theta^*} \in \mathcal{F}$ are required as the initial guess. In our implementation, the method presented in~\cite{fulton2019latent} for the supervised setting and the method in ~\cite{sharp2023data} for the unsupervised setting is applied to first minimize $\mathcal{L}_C$ and find the image ${\rm{Im}}(\bm{f}_{\theta})$ of the subspace mapping $\bm{f}_{\theta}$. Then the landscape of the subspace is optimized by minimizing both construction and Lipschitz loss together as
\begin{equation}
\label{form-unconstrained-lipschitz-optimization}
\min_\theta \left[ \mathcal{L}_{\rm{C}}(\theta) + \lambda_{\rm{LS}}\mathcal{L}_{LS}(\bm{f}_\theta) \right],
\end{equation}
where $\lambda_{\rm{LS}}$ controls the trade-off between the constraint compliance and the Lipschitz optimization, which can balance the simulation quality and convergence speed of the simulation.

Training using combined loss \cref{form-L-LS-E} requires observations $\mathcal{Z}$ sampled in subspace. For the supervised setting \cref{form-vanilla-supervised-construction}, we obtain the observations $\mathcal{Z}$ from sampling on the dataset $\mathcal{Q}$ and pulling the samples back to the subspace coordinates through the encoder $\bm{g}_\theta$. For the unsupervised setting \cref{form-vanilla-unsupervised-construction}, the pull-back distribution is set as the standard Gaussian distribution $\mathcal{N}$. Note that for dynamic simulation, sampling on the inertia term will greatly increase the variation on $\mathcal{L}_{LS}$ estimations and therefore cannot provide a meaningful signal for the training. Thankfully the major nonlinearity in the optimization problem comes from the elasticity term $P(\mathbf{q})$, in this work we only apply for Lipschitz optimization on the elasticity term (i.e., compute for $\nabla_{\bm{z}}^2 P(\bm{z})$), with discussion presented in Sec.~\ref{sec-limitation}.

\subsection{Cubature Approximation}
\label{sec-cubature-method}

\begin{figure}[t]
  \centering
  \includegraphics[width=1\linewidth]{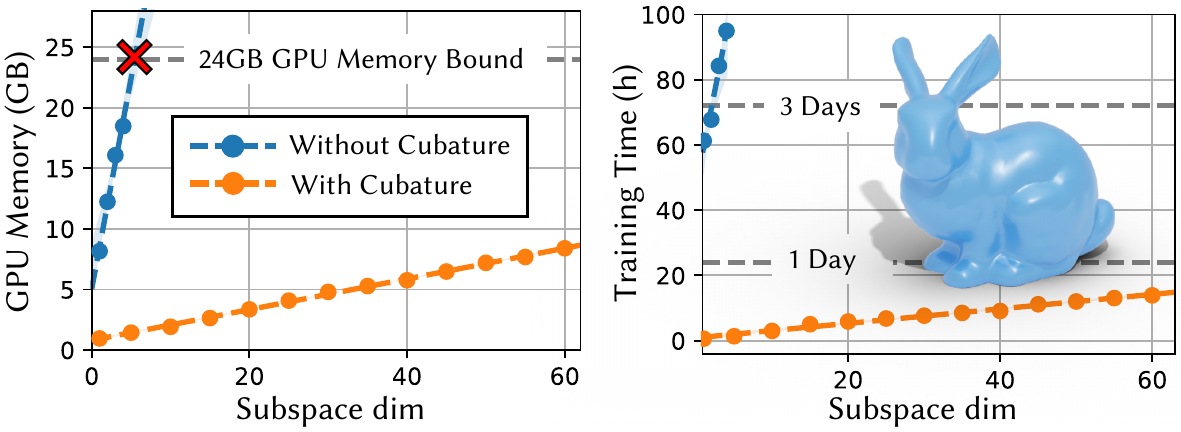}
  \caption{Comparison of the relationship between subspace dimension and training cost without and with the cubature approximation. The data points are collected on the bunny problem with $44k$ DOFs and $53k$ tetrahedrons using a RTX3090. $300$ cubatures are used in this example. }
  \label{fig-cubature-training-acceleration}
\end{figure}

When adding Lipschitz Loss into the training process, two additional passes of backpropagation are required for the computation of each row or column of the Hessian matrix $\nabla_{\bm{z}}^2 P(\bm{z})$, resulting in a great increase in memory usage. Considering the network scale as $N$, batch size as $B$, number of elements as $E$, subspace dimension as $r$, and potential-specific coefficients as $p$,
with Lipschitz loss, the overall storage cost of training is $\bm{O}(2rB(r+N+n+pE))$. 
Compared with the cost of $\bm{O}(B(r+N+n))$ for conventional supervised method using only $
\mathcal{L}_{C}$, adding $\mathcal{L}_{LS}$ causes severe memory shortages when training the network on high-resolution meshes. An example is presented in Fig.~\ref{fig-cubature-training-acceleration}, directly optimize $\mathcal{L}_{LS}$ on the bunny model with 52k tetrahedrons with a very small  \revised{neural} subspace (less than 10) already reach the GPU memory bound at 24 GB. To cut down the memory cost and accelerate the training, we leverage the cubature method ~\cite{von2013efficient} which is often used for runtime acceleration in reduced-order simulations to make a workaround.

As discussed in Sec.~\ref{sec-training-details}, we focus on optimizing Lipschitz loss on elastic potential $P$ of the simulation, and cubature methods ~\cite{an2008optimizing, von2013efficient, trusty2023subspace, yang2015expediting} have been widely studied to approximate the gradients of $P$ with respect to the subspace coordinates $\bm{z}$ only using a small subset $\mathcal{S}$ of all elements $\mathcal{E}$:
\begin{equation}
\label{form-cubature-gradients}
\nabla_{\bm{z}}P=\sum_{e\in\mathcal{E}}\frac{\partial P_e}{\partial \bm{q}}\frac{\partial \bm{f}_{\theta}}{\partial \bm{z}}\approx\sum_{e\in\mathcal{S}}w_e\frac{\partial P_e}{\partial \bm{q}}\frac{\partial \bm{f}_{\theta}}{\partial \bm{z}},
\end{equation}
where $P_e$ is the elastic potential of element $e$ and $w_e$ is a non-negative weight for element $e$. 
It's generally the case that the subset size $S$ is selected to be within an order of magnitude of the dimension of the subspace (i.e., make $S \sim r$ and $r \ll E$)~\cite{an2008optimizing}, and in this way, the computational complexity, as well as storage cost of $\nabla_{\bm{z}}P$ can be greatly reduced from $\bm{O}(r+N+n+pE)$ to $\bm{O}(r+N+pS)$. 
We apply a similar strategy to estimate the subspace Hessian by first approximating the potential objective as $\tilde{P}=\sum_{e\in\mathcal{S}}w_e  P_e$ and then using $\nabla_{\bm{z}}^2\tilde{P}$ as a Hessian estimation in the Lipschitz loss \cref{form-L-LS-E}, then the storage cost of the Lipschitz loss can be lowered to $\bm{O}(4rB(r+N+pS)) \sim \bm{O}(BprS))$. It is worth mentioning that the training process is also accelerated due to the reduced computational cost.
\revised{We note that different from existing work utilizing the cubature approximation to estimate subspace gradients during runtime (e.g. \cite{fulton2019latent, shen2021high, trusty2023subspace}), we use cubatures to approximate the subspace Hessian, which facilitates the training process. Please refer to the supplementary document for further discussion on using cubatures during runtime.}
Since the construction of the cubature subset and weights requires a deterministic subspace mapping ~\cite{von2013efficient}, the initial subspace obtained by only optimizing $\mathcal{L}_C$ is used to generate the cubatures.
As presented in Table.~\ref{table-problems-statistics}, we empirically show that even with cubature approximation, our Lipschitz optimization still achieves simulation speedup. The selection of hyperparameter S is discussed in Sec.~\ref{sec-cubature-number-selection}.

\begin{table*}
\caption{Parameters and results of experiments evaluated in this work. $n$: number of full-space DOFs; $E$: number of mesh elements; $r$: the subspace dimension used; $S$: number of cubatures used during \textit{training}; ${\rm{Lip}}[\nabla_{\bm{z}}^2 P]$: the estimated Lipschitz constant of the subspace potential Hessian.}
\label{table-problems-statistics}
\begin{tabular}{c|c|c|c|c|c|c|c|c|c|c|c|c}
    \hline
    \multirow{2}{*}{Example} & \multirow{2}{*}{Figure} & \multirow{2}{*}{$n$} & \multirow{2}{*}{$E$} & \multirow{2}{*}{$r$} & \multirow{2}{*}{$S$} & \multicolumn{2}{c|}{Training Time (h)} & \multicolumn{2}{c|}{${\rm{Lip}}[\nabla_{\bm{z}}^2 P]$} & \multicolumn{3}{|c}{Simulation step time (ms)}  \\
    \cline{7-13}
    & & & & & & Ours & Vanilla & Ours & Vanilla & Ours & Vanilla & Full \\
    \hline 
    Bistable$^{\dagger\ast}$ & Fig.~\ref{fig-unsupervised-tests}(a) & 1.6K & 1K & 8 & - & 1.0 (2.3x) & 0.4 & 20.2 & 30.9 (1.53x) & 2.1 & 3.9 (1.87x) & 82.5 (39.08x) \\
    \hline
    Cloth$^{\dagger\ast}$ & Fig.~\ref{fig-unsupervised-tests}(b) & 7K & 4.5K & 8 & - & 9.0 (10.0x) & 0.9 & 3.8 & 9.0 (2.78x) & 16.3 & 23.1 (1.42x) & 73.1 (4.48x)\\
    \hline
    Twist Bar$^{\ddagger\ast\ast}$ & Fig.~\ref{fig-compress-bar-quality} & 4.7K & 3.1K & 10 & 300 & 2.6 (4.8x) & 0.6 & 28.4 & 299.7 (10.57x) & 4.4 & 12.3 (2.78x) & 115.6 (26.22x) \\
    \hline
    Dinosaur$^{\ddagger\ast}$ & Fig.~\ref{fig-teaser} & 23K & 29K & 40 & 550 & 18.3 (4.4x) & 4.2 & 0.3 & 215.8 (644.22x) & 9.1 & 62.6 (6.83x) & 1182.4 (129.93x)\\
    \hline
    Elephant$^{\ddagger\ast}$ & Fig.~\ref{fig-image-quality} & 38K & 62K & 65 & 400 & 22.3 (4.8x) & 4.7 & 1.3 & 117.5 (92.48x) & 14.1 & 38.9 (2.76x) & 464.2 (32.85x)\\
    \hline
    Bunny$^{\ddagger\ast}$ & Fig.~\ref{fig-bunny} & 44K & 53K & 40 & 300 & 10.0 (5.8x) & 1.7 & 0.3 & 42.6 (125.24x) & 4.5 & 21.0 (4.72x) & 687.2 (154.43x)\\
    \hline
\end{tabular}
\begin{flushleft}\small
 $^\dagger$unsupervised setting; $^\ddagger$supervised setting; $^\ast$dynamic simulation; $^{\ast\ast}$quasi-static simulation. For dynamic simulations, we use a timestep of $50$~ms (i.e., 20 fps).
\end{flushleft}
\end{table*}

\begin{figure}[t]
  \centering
  \includegraphics[width=1\linewidth]{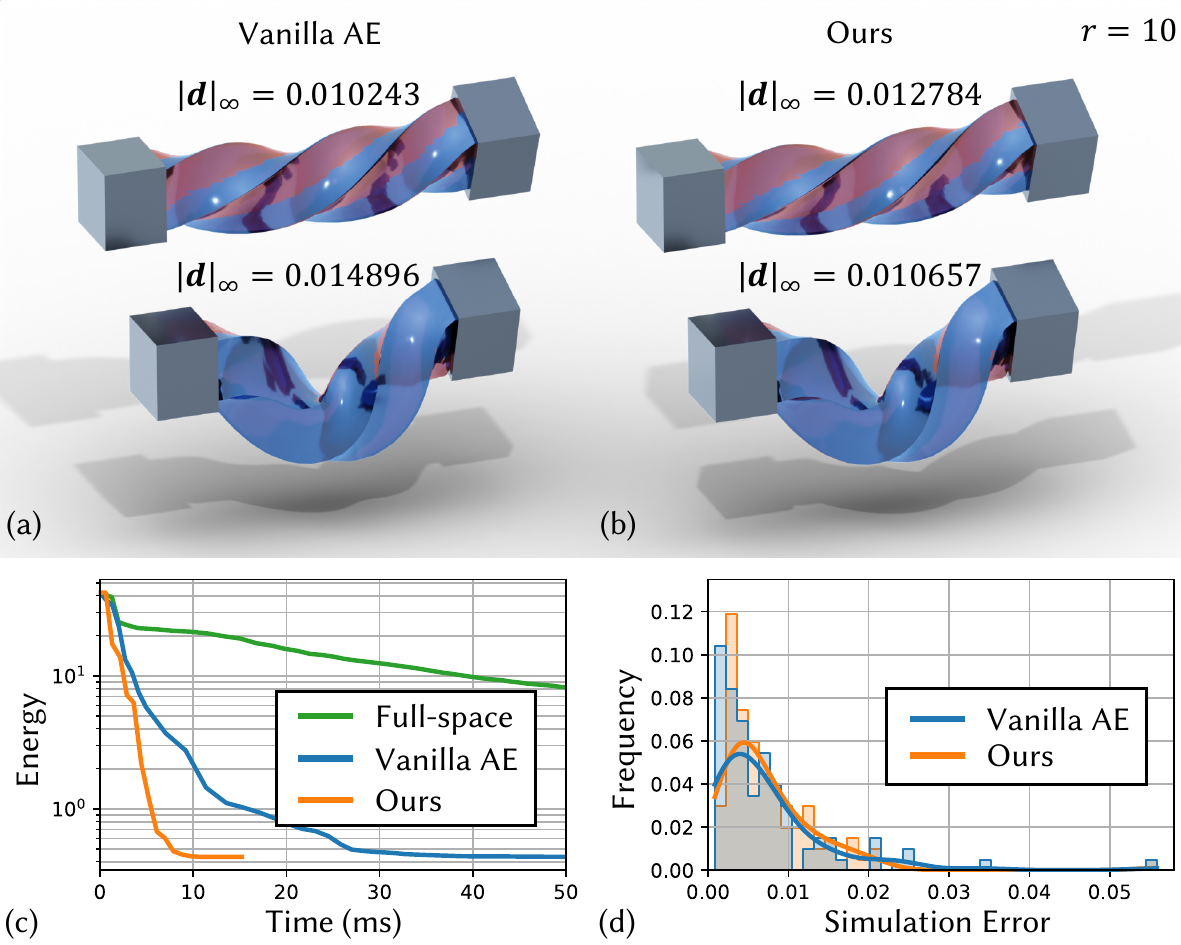}
  \caption{ Performance on a compressed and twisted bar. The simulation results using subspaces are rendered in blue, while full-space simulation results are overlaid in purple as reference shapes. (a) Results of vanilla neural subspace. (b) Results of our method with optimized Lipschitz energy. (c) Comparison of the convergence speed in the simulation when converging to a similar termination energy. (d) Simulation error distribution with full-space simulation as the reference. The solid lines are Kernel Density Estimation (KDE) plots that visualize the estimated probability density curves of the simulation error. }
  \label{fig-compress-bar-quality}
\end{figure}

\section{Result and Discussion}

We test the performance of the proposed Lipschitz optimization method in various physical systems and demonstrate that our method can effectively improve the Lipschitz constant of the subspace potential Hessian, resulting in simulation speedups. Furthermore, our method conserves subspace quality, thus achieving similar configuration manifold coverage and comparable simulation quality (please refer to our supplemental video for all results). The physical systems tested include two unsupervised settings and four supervised settings, each with a comparison with subspace learning that minimizes only $\mathcal{L}_C$ (denoted as "Vanilla"). Details of the performance in terms of training time, Lipschitz constants of subspace Hessians, and mean simulation speed can be found in Table~\ref{table-problems-statistics}.

Our implementation is based on JAX~\cite{jax2018github}. The Hessian matrix is computed using autodiff. For the optimizer used to solve \cref{form-reduced-space-time-integration}, we employ the L-BFGS with line search for all simulations. Tests using the projective Newton's method as the timestepping optimizer are also presented in the supplementary document. Four termination conditions are applied: 1) The gradient norm of the objective falls below $\epsilon = 10^{-5}$ (i.e., converged); 2) L-BFGS encounters a saddle point; 3) Maximum line search iterations (set as $128$) reached; 4) Maximum L-BFGS iterations (set as $1024$) reached. A discussion about the selection of the convergence condition can be found in the supplementary document. All training processes and experiments were conducted on an NVIDIA RTX 3090 Graphics card.

\begin{figure}[t]
  \centering
  \includegraphics[width=1\linewidth]{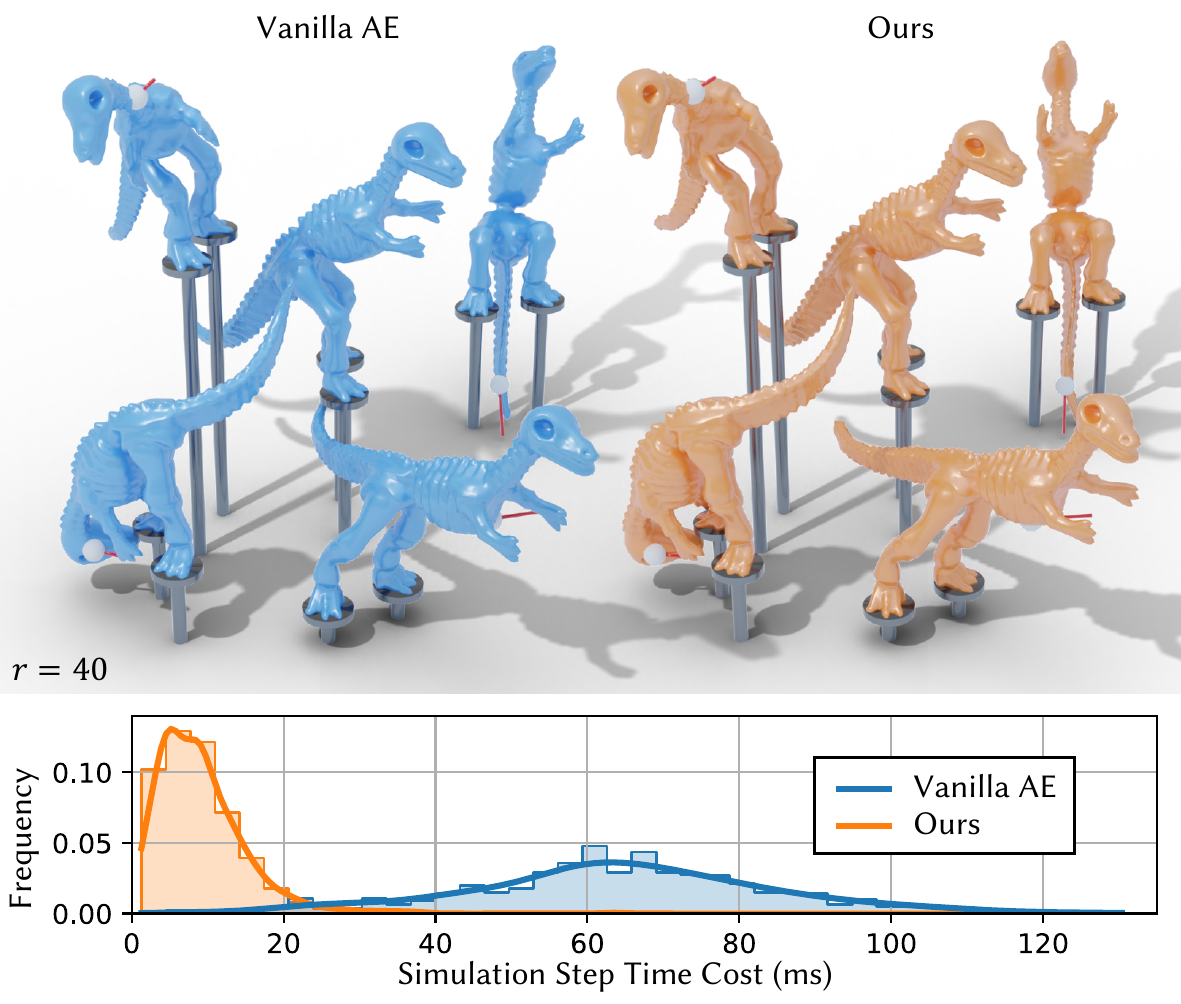}
  \caption{ Our method and the vanilla subspace construction produce complex deformations on the dinosaur mesh by applying interactions. We also report the simulation time cost distribution of the two methods. }
  \label{fig-complex-deformation}
\end{figure}

\begin{figure}[t]
  \centering
  \includegraphics[width=1\linewidth]{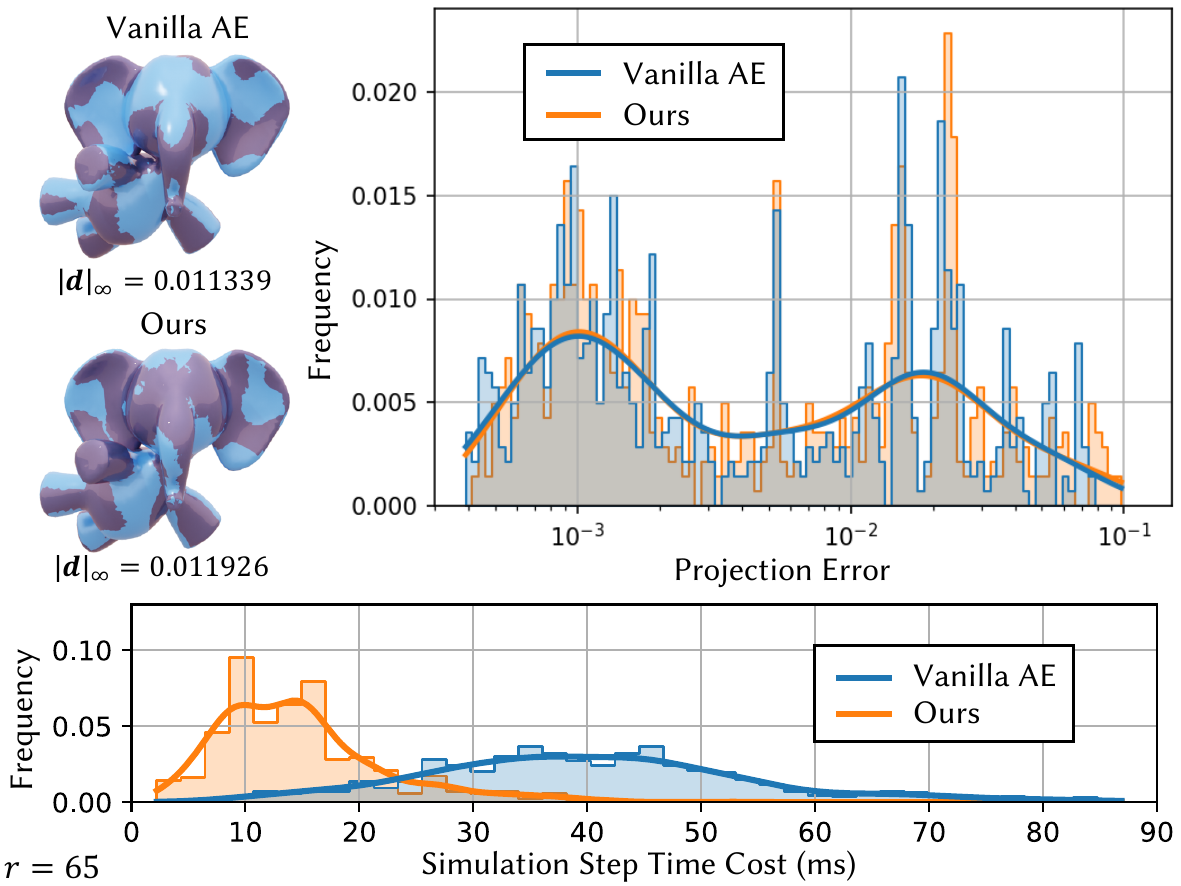}
  \caption{ For the elephant example, we project full-space simulation's state samples to the configuration manifold induced by different subspace constructions. The projected shapes are rendered in blue and the source shapes are rendered in purple. We also report the projection error and simulation time distribution.  }
  \label{fig-image-quality}
\end{figure}

\subsection{Evaluation on Supervised Settings}
For all four examples, we use the stable neo-Hookean material model~\cite{smith2018stable} to define the elastic potential energy. We prepare the training sets that typically contain $\sim$$3000$ frames by randomly sampling the interactions and moving constraints involved in these examples and running full-space simulations. For all models, we train the model with $\mathcal{L}_{C}$ and $\mathcal{L}_{LS}$ together around 20k epochs. When evaluating ${\rm{Lip}}[\nabla_{\bm{z}}^2 P]$ presented in Table~\ref{table-problems-statistics}, we sample from the pull-back distribution and use \cref{form-H-Lipschitz-estimation} for the estimation. Since the pull-back distribution in supervised settings is not explicit, its sampling is realized by sampling the configuration distribution (i.e., test-set full-space simulation snapshots) and then projecting to the subspace.

\begin{figure}[t]
  \centering
  \includegraphics[width=1\linewidth]{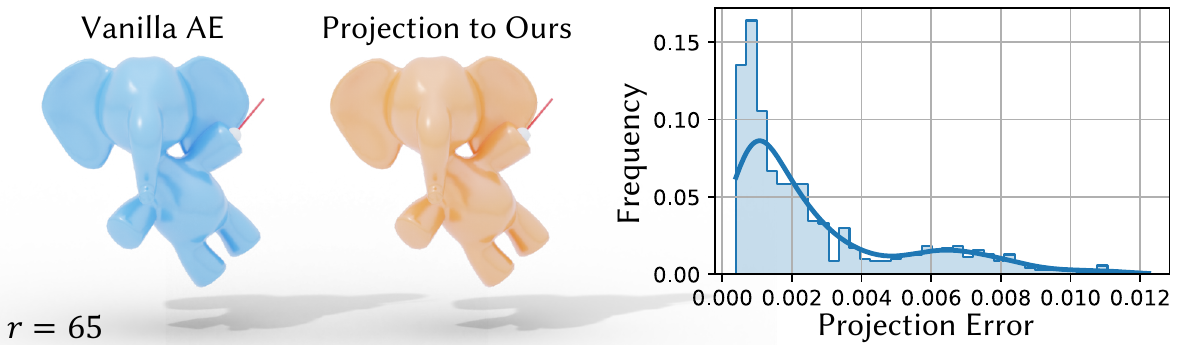}
  \caption{ We project state samples from the vanilla neural subspace construction onto the configuration manifold induced by our subspace construction. The low projection error in this example indicates that the two mappings' images are similar.  }
  \label{fig-image-similarity}
\end{figure}

\subsubsection{Twist Bar with Quasi-static Simulation}

As illustrated in Fig.~\ref{fig-compress-bar-quality}, positional constraints are applied on the sides of the bar, making the bar twisted and compressed to varying degrees. It can be seen from the convergence curve in Fig.~\ref{fig-compress-bar-quality}(c) that by using our Lipschitz optimization method, the convergence speed of the shown neural reduced-order simulation is accelerated by $\sim$3 times to converge to the same level of potential energy. Meanwhile, by comparing with the ground-truth result obtained from the full space simulation, The error distribution presented by kernel density in Fig.~\ref{fig-compress-bar-quality}(d) of our method is analogous to that of the vanilla neural subspace construction (i.e., only minimize $\mathcal{L}_C$), showing a comparable simulation quality of our method.

\begin{figure}[t]
  \centering
  \includegraphics[width=1\linewidth]{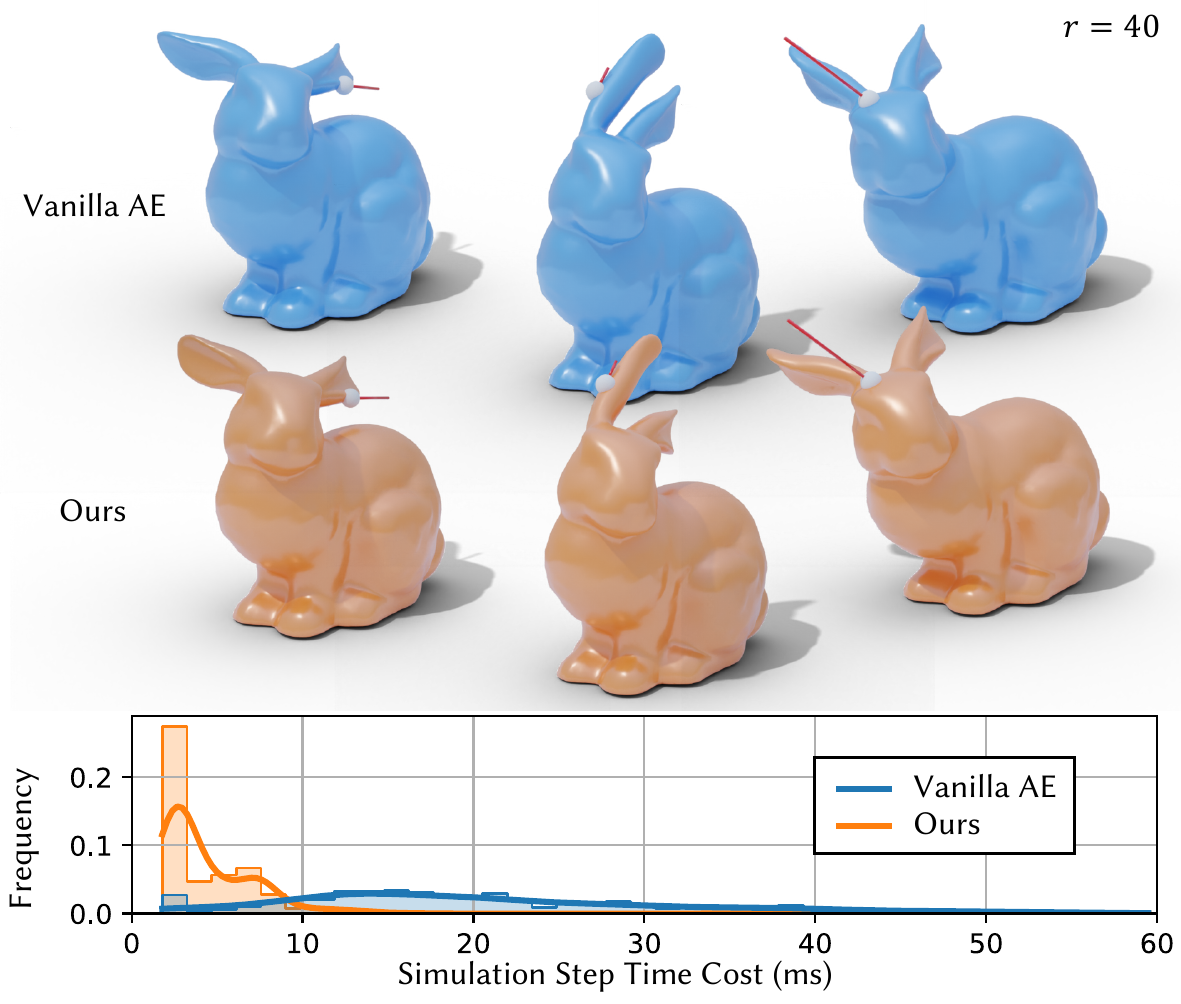}
  \caption{Result of the bunny example simulated with interactions. Compared with Vanilla method, the neural subspace with optimized Lipschitz characteristics can obtain a faster speed while keeping the quality of simulation.
  }
  \label{fig-bunny}
\end{figure}

\subsubsection{Interactive Simulation with Dynamics} 
\label{section-supervised-dynamics-results}

The computational efficiency of our method is demonstrated on three examples with dynamic simulations: dinosaur (Fig.~\ref{fig-complex-deformation}), elephant (Fig.~\ref{fig-image-quality}), and bunny model (Fig.~\ref{fig-bunny}). 
For these examples, the average simulation step time for 500 frames in the test set has decreased by factors of 6.83, 2.76, and 4.72, showcasing the great performance of our method in optimizing the landscape of the subspace objective. 
On the other hand, by projecting the computed states onto the configuration manifold, we can evaluate the projection error as the point-to-point distance between two deformed shapes. As illustrated in Fig.~\ref{fig-image-quality}, the distribution of projection error of the elephant example between the Vanilla method and ours to the ground truth (full space samples) is similar. Meanwhile, the projection error between our method and the Vanilla subspace result is concentrated at 1‰ (1 mm; the diameter of the bounding box of the mesh we use is 1 m) — see Fig.~\ref{fig-image-similarity}. This shows that we successfully optimized the Lipschitz regularizers while maintaining the image of the neural subspace (i.e., preserving the configuration manifold), resulting in a similar simulation quality as conventional neural-network-based methods.

\subsection{Evaluation on Unsupervised Settings}
For unsupervised settings, the network can automatically find a condensed subspace. For both models (the bistable bar and the cloth), we select the subspace dimension as 8. The neural subspace is constructed by first training around one million iterations with only $\mathcal{L}_{C}$ (as the Vanilla model) then followed by 200k iterations adding $\mathcal{L}_{LS}$. 

For both examples, the 2D version of the stable neo-Hookean model \cite{smith2018stable} is used. In particular, for the cloth simulation, a hinge-based bending model~\cite{grinspun2003discrete} is further added for bending elasticity. Contact with rigid bodies is handled by detecting vertex contacts using the rigid body's signed distance function (SDF) and applying penalty forces for collision response. The samples for estimating ${\rm{Lip}}[\nabla_{\bm{z}}^2 P]$ for unsupervised settings are directly obtained by Gaussian sampling in the subspace.

\begin{figure}[t]
  \centering
  \includegraphics[width=1\linewidth]{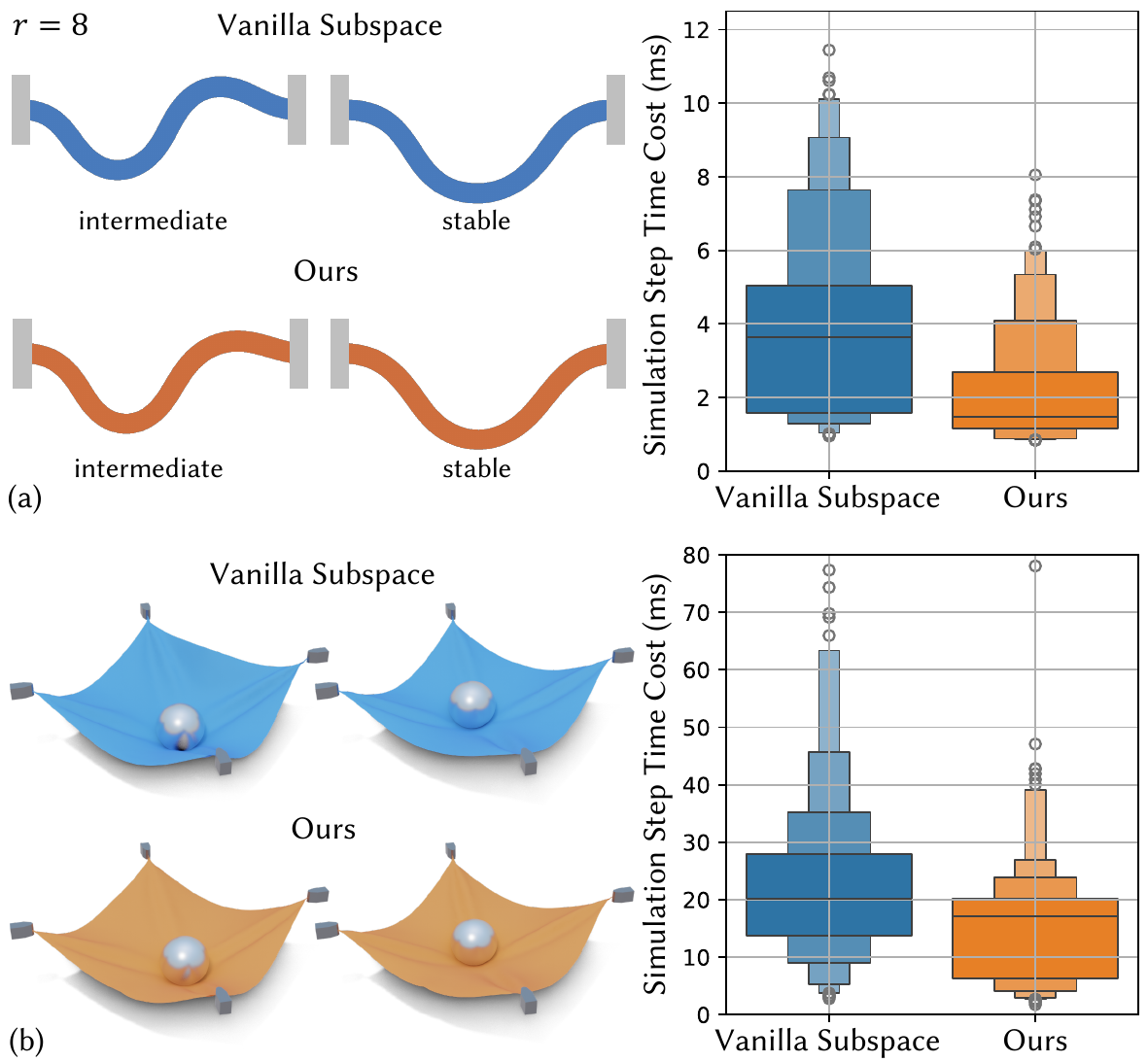}
  \caption{Comparison of performance (simulation speed and quality) on (a) a bistable bar, and (b) a cloth simulation with interactions, where the subspaces are constructed by unsupervised learning. The box plots on the right present the stepping time distribution, with the box height representing the interquartile range and the width indicating the number of data points within the corresponding range. For the vanilla subspaces, a few outliers ($\sim$$1\%$ of the data) were omitted to avoid over-compression of the plot.
  }
  \label{fig-unsupervised-tests}
\end{figure}

The result for the 2D compressed bar with dynamics is presented in Fig.~\ref{fig-unsupervised-tests}(a). It can be seen that our method performs $1.87\times$ faster compared to the Vanilla method. Meanwhile, both methods achieve similar intermediate and stable states. The other tested system is illustrated in Fig.~\ref{fig-unsupervised-tests}(b), where we show that by treating collision handling as a part of the potential energy, our method can speed up simulations involving contacts. In this example, a cloth drape interacts with a rigid ball through energy-based collisions. Our method achieves an acceleration rate of $1.42\times$ in this scenario, with comparable simulation quality.

It can be noticed that the acceleration performance of unsupervised settings is not as good as that of supervised ones. This is mainly because the subspace dimension of the unsupervised setting is very small, which leads to a manifold with less room to perform Lipschitz optimization. When increasing the subspace dimension, we found it very challenging to find appropriate hyperparameters to make the training stable. A detailed discussion on subspace dimension and acceleration performance is presented in the next section.

\subsection{Comparison on Hyperparameters Selection}

\begin{figure}[t]
  \centering
  \includegraphics[width=1\linewidth]{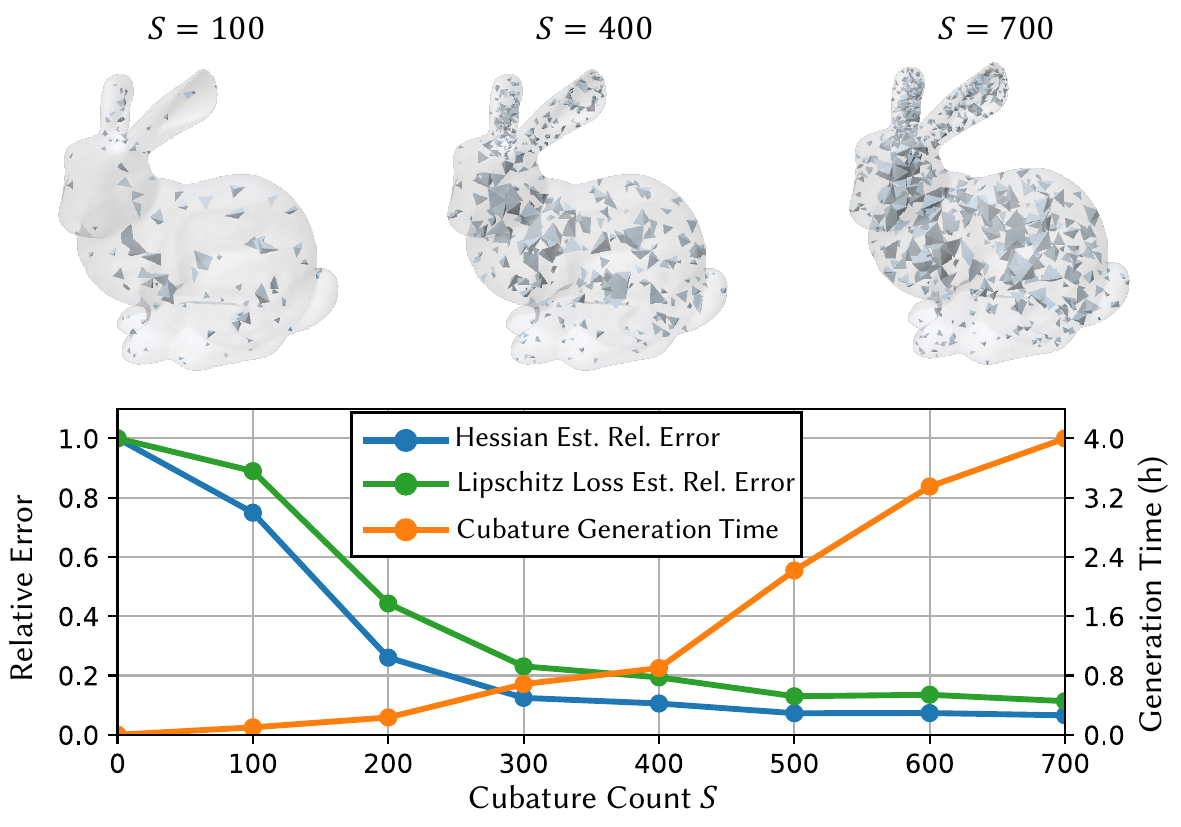}
  \caption{For the Bunny problem, we generate a series of different numbers of cubatures and compare their estimation accuracy for the Lipschitz loss and the hessian, as well as the generation time consumed.}
  \label{fig-cubature-count-tradeoff}
\end{figure}

\subsubsection{Cubature Number Selection}
\label{sec-cubature-number-selection}
As illustrated in Fig.~\ref{fig-cubature-training-acceleration}, using the cubature method \cite{von2013efficient} can greatly reduce the training cost of the Lipschitz regularizer. We carefully select the number of cubatures as an important hyperparameter in training. As presented in Fig.~\ref{fig-cubature-count-tradeoff}, for the bunny model, the number of cubatures determines the estimation accuracy for the Hessian as well as for the Lipschitz loss. However, an excess number of cubatures results in higher training time and significantly more time spent on the generation of the cubature set itself. When the number of cubatures exceeds 400, the generation time increases significantly, but the decrease in estimation error slows down. Based on numerical experiments, we found that a Lipschitz loss estimation error of around 20\% is sufficient to obtain a considerable simulation speedup. Therefore, we choose numbers of cubatures around 400 to balance estimation accuracy and computational speed.

\subsubsection{Influence of Subspace Dimension on Simulation Quality and Speed}
\begin{figure}[t]
  \centering
  \includegraphics[width=1\linewidth]{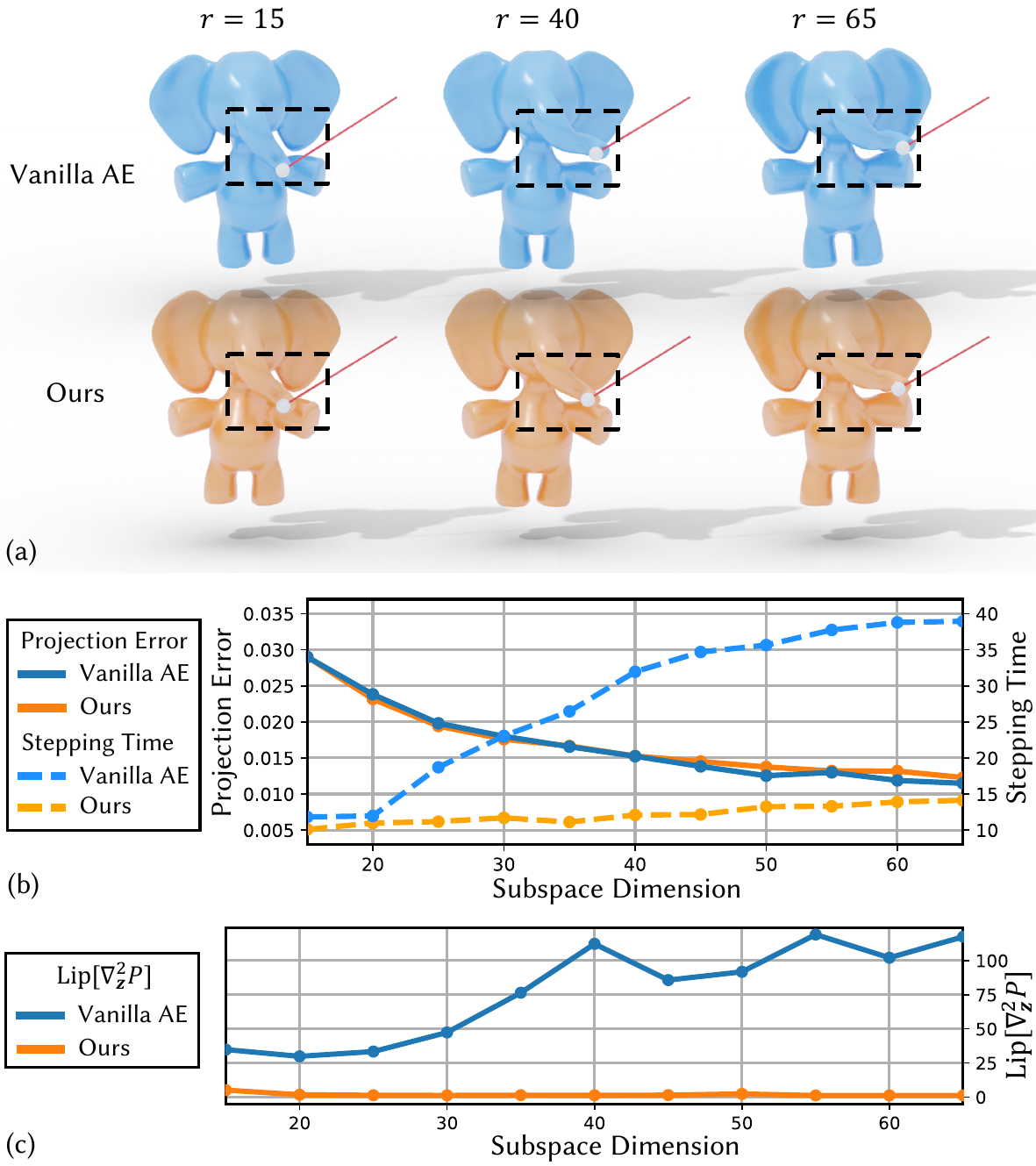}
  \caption{ We construct subspaces of different dimensions for the Elephant problem and compare their simulation quality and speed. (a) Qualitative comparison on simulation quality by applying a fixed interaction to the trunk. (b) Qualitative comparison on simulation quality and speed by mean error of full-space simulation states projections and mean simulation step time cost. (c) Qualitative comparison on the Lipschitz constants of the subspace Hessian. }
  \label{fig-dim-analysis}
\end{figure}

The subspace dimension is an important hyperparameter for all reduced-order simulation methods as it can directly affect the quality of the subspace simulation, and we also find that this value affects the speedup ability of our method. We constructed subspaces of different subspace dimensions $r$ for the Elephant simulation, and the result is presented in Fig.~\ref{fig-dim-analysis}. We found that with the increase in the subspace dimension, the acceleration performance of our method improves as the simulation time increases significantly for the vanilla neural-based method. Meanwhile, for all sizes of subspaces, our method consistently maintains similar simulation quality to the vanilla neural subspace. This demonstrates that by using our method with Lipschitz optimization, we can improve the simulation quality by increasing the subspace dimension at a small cost in simulation time.

We also empirically find that the Lipschitz constants are successfully lowered compared to the vanilla method in different dimension settings, as shown in Fig.~\ref{fig-dim-analysis}(c). We note that our sampling performs on the configuration manifold $\mathcal{M}$. Its dimension for a certain setting is determined a priori, independent of the selection of the subspace dimension $r$ and the learned subspace mapping $\bm{f}_{\theta}$. Therefore, our method does not suffer from sampling issues associated with high subspace dimensions. 

\subsection{Limitation}
\label{sec-limitation}

In this work, Lipschitz optimization is only applied to the elastic potential term of \cref{form-reduced-space-time-integration}. Since the nonlinear mapping $\bm{f}(\bm{z})$ is also involved in the inertia term, this may lower the convergence speed of the simulation involving dynamics. Considering that the inertia term is in quadratic form, the Hessian Lipschitz of the inertia term can be optimized by minimizing or bounding the Lipschitz constant of the network's input-output Jacobian~\cite{gulrajani2017improved, gouk2021regularisation, liu2022learning}. This is a promising direction for future work to further accelerate the simulation with dynamics.

Another limitation of our method is the extended training time introduced by incorporating Lipschitz optimization into the pipeline. As shown in Table~\ref{table-problems-statistics}, even with cubature acceleration, the training time is still increased by approximately five times compared to the conventional method. This issue can be addressed by employing fast approximate methods to estimate Lipschitz energy.

\section{Conclusion}

We present a method that incorporates Lipschitz optimization to accelerate the simulation of deformables using a \revised{neural} reduced-order solver. We show that by using our method the landscape of the simulation objective in the subspace is optimized, resulting in reduced optimizing time of the solver. Cubature approximation is employed to facilitate a successful training process by reducing the usage of GPU memory and training time. Our work achieves acceleration factors ranging from 1.42 to 6.83 across various cases involving complex deformations (e.g., twisting, bending, and interactive deformation) and works effectively for both supervised and unsupervised settings.

\begin{acks}
The authors would like to thank the anonymous reviewers for their valuable comments. 

The project is supported by National Key Research and Development Program of China (2022YFE0112200), National Natural Science Foundation of China (U21A20520, 62325204), Key-Area Research and Development Program of Guangzhou City (202206030009), HKSAR RGC Early Career Scheme (ECS) (CUHK/24204924).
\end{acks}

\bibliographystyle{ACM-Reference-Format}
\bibliography{reference}


\appendix
\begin{table}[b]
\vspace{-0.2cm}
\caption{Comparisons of selections for the convergence condition's threshold $\epsilon$ using the dinosaur example. }
\label{table-convergence-condition-comparison}
\small
\begin{tabular}{c|c|cccc|c|c}
\hline
\multirow{2}{*}{$\epsilon$}   & \multirow{2}{*}{Subspace} & \multicolumn{4}{c|}{Exit Condition Triggered} & \multirow{2}{*}{$\overline{\|\bm{g}^{\rm{exit}}\|}_{\infty}$} & \multirow{2}{*}{$\overline{t}$} \\ \cline{3-6}
                           &                           & \multicolumn{1}{c|}{I} & \multicolumn{1}{c|}{II} & \multicolumn{1}{c|}{III} & \uppercase\expandafter{IV} &                                                                                      \\ \hline
\multirow{2}{*}{$10^{-4}$} & Ours                      & \multicolumn{1}{c|}{100.0\%}   & \multicolumn{1}{c|}{0.0\%}        & \multicolumn{1}{c|}{0.0\%}            & 0.0\%    &  $7.2\times10^{-5}$  & 3.7                                                                                 \\ \cline{2-8} 
                           & Vanilla                   & \multicolumn{1}{c|}{99.5\%}    & \multicolumn{1}{c|}{0.0\%}        & \multicolumn{1}{c|}{0.5\%}            & 0.0\%    &  $8.8\times10^{-5}$  & 31.3                                                                        \\ \hline
\multirow{2}{*}{$10^{-5}$} & Ours                      & \multicolumn{1}{c|}{98.1\%}    & \multicolumn{1}{c|}{1.9\%}        & \multicolumn{1}{c|}{0.0\%}            & 0.0\%    &  $8.3\times10^{-6}$  & 9.1                                                                                  \\ \cline{2-8} 
                           & Vanilla                   & \multicolumn{1}{c|}{32.1\%}    & \multicolumn{1}{c|}{67.4\%}       & \multicolumn{1}{c|}{0.4\%}            & 0.0\%    &  $5.1\times10^{-5}$  & 62.6                                                                        \\ \hline
\multirow{2}{*}{$10^{-6}$} & Ours                      & \multicolumn{1}{c|}{89.3\%}    & \multicolumn{1}{c|}{10.7\%}       & \multicolumn{1}{c|}{0.0\%}            & 0.0\%    &  $1.8\times10^{-6}$  & 28.2                                                                                \\ \cline{2-8} 
                           & Vanilla                   & \multicolumn{1}{c|}{0.0\%}     & \multicolumn{1}{c|}{96.4\%}       & \multicolumn{1}{c|}{3.6\%}            & 0.0\%    &  $5.0\times10^{-5}$  & 63.4                                                                    \\ \hline
\end{tabular}
\begin{flushleft}\small
I: Converged nominally - the gradient norm of the objective falls below $\epsilon$; \\ II: L-BFGS encounters a saddle point; \\ III: Maximum line search iterations (set as $128$) reached; \\ IV: Maximum L-BFGS iterations (set as $1024$) reached.
\end{flushleft}
\end{table}

\section{Supplementary Document}

\subsection{Discussion on Convergence Condition Selection}
When using iterative solvers such as L-BFGS, the time cost of solving \cref{form-reduced-space-time-integration} is largely influenced by the selection of the convergence condition. We use the gradient norm $\|\bm{g}\|_{\infty}$ of the objective $E$ as the criterion. The optimization is treated as converged if the gradient norm $\|\bm{g}\|_{\infty}$ falls below a threshold $\epsilon$. For all results reported in this work except in this section, we take $\epsilon=10^{-5}$ to keep the same termination conditions as in~\cite{sharp2023data}.

We further discuss here the influence of the selection of convergence conditions. As shown in the Table~\ref{table-convergence-condition-comparison}, we collect the triggered exit conditions statistics, mean exit gradient norm $\overline{\|\bm{g}^{\rm{exit}}\|}_{\infty}$, and mean simulation step time cost (ms) $\overline{t}$ for the dinosaur example using different $\epsilon$ values. We find that with $\epsilon=10^{-4}$, both our method and vanilla subspace construction can converge nominally for nearly all of the timesteps, and the acceleration rate reaches 8.43x. By making the convergence condition stricter (i.e. taking $\epsilon=10^{-5}$), the convergence condition becomes harder to be triggered for the vanilla method, while our method is still able to iterate to a solution with a sufficiently small gradient norm (i.e., $\overline{\|\bm{g}^{\rm{exit}}\|}_{\infty} < \epsilon$). Not being able to reach a small gradient norm means the iterations of the vanilla method stop earlier at solutions of lower quality. With a further lowered threshold $\epsilon=10^{-6}$, the vanilla method fails to trigger the convergence condition in any of the timesteps, resulting in a mean timestepping cost similar to that under $\epsilon=10^{-5}$. It is also worth mentioning that as $\epsilon$ decreases from $10^{-5}$ to $10^{-6}$, the mean exit gradient norm of the vanilla method does not show a sufficient decrease and remains well above the designated convergence threshold. In contrast, our method shows better convergence thanks to the optimized subspace landscape. 

\begin{table}[t]
\caption{Performance and Lipschitz constants comparisons for different orders' Lipschitz losses.}
\label{table-general-lipschitz-loss}
\small
\begin{tabular}{c|c|c|c|c}
\hline
Lipschitz Loss            & \begin{tabular}[c]{@{}c@{}}Simulation\\ step time (ms)\end{tabular} & ${\rm{Lip}}[P]$ & ${\rm{Lip}}[\nabla_{\bm{z}} P]$ & ${\rm{Lip}}[\nabla_{\bm{z}}^2 P]$ \\ \hline
Vanilla                   & 21.0                                                                & 1.84            & 9.59                            & 42.58                             \\ \hline
$\mathcal{L}_{\rm{LS},0}$ & 25.3                                                                & 1.18            & 18.37                           & 108.41                            \\ \hline
$\mathcal{L}_{\rm{LS},1}$ & 13.0                                                                & 0.49            & 1.79                            & 6.40                              \\ \hline
$\mathcal{L}_{\rm{LS},2}$ & 4.5                                                                 & 0.32            & 0.17                            & 0.34                              \\ \hline
\end{tabular}
\vspace{-0.2cm}
\end{table}

\subsection{Lipschitz Energies of different orders}

In this work, we focus on the Lipschitz regularization of subspace Hessians (i.e., \cref{form-L-LS-E}), which, in other words, is a second-order Lipschitz optimization. One can also generalize the Lipschitz loss to arbitrary order $o$:
\begin{equation}
\label{form-L-LS-E-general}
\mathcal{L}_{\rm{LS},o} = \mathbb{E}_{\bm{z}_1, \bm{z}_2 \overset{\mathrm{iid}}{\sim} \Pi_\theta(\bm{z})}\left[ 
\frac{\left\| \frac{\partial^o P}{\partial \bm{z}^o}(\bm{z}_1) - \frac{\partial^o}{\partial \bm{z}^o}(\bm{z}_2) \right\|^2 } {||\bm{z}_1 - \bm{z}_2||^2} 
\right],
\end{equation}
where our Lipschitz loss ($\mathcal{L}_{LS}$ defined by~\cref{form-L-LS-E} in the main content) becomes a special case $\mathcal{L}_{\rm{LS},2}$. For the bunny example, we additionally test the acceleration rate of different orders of Lipschitz regularization, and the results are reported in Table~\ref{table-general-lipschitz-loss}. Note that we only test schemes where $o \leq 2$ as losses with higher orders result in overly high training costs (see Sec.~\ref{sec-cubature-method}). We find that accelerations produced by lower-order Lipschitz losses lag significantly behind those of the 2nd-order one (i.e. \cref{form-L-LS-E}). This aligns with our theoretical analysis in Sec.~\ref{sec-convergence-discussion}. We also find that optimizing the 2nd-order Lipschitz loss also shows a decrease in lower-order Lipschitz constants.

\subsection{Tests with Projective Newton's method}
Besides the L-BFGS, we test the projective Newton's method with line search as the optimizer for \cref{form-reduced-space-time-integration}. The performance statistics for each example are shown in Table~\ref{table-projective-newton-statistics}. Our method simultaneously shows acceleration over vanilla subspace constructions when using the different optimizer. For examples with large subspace dimensions, both our subspace constructions and vanilla subspace constructions experience slower timestepping compared to that using L-BFGS. This is due to the expensive subspace Hessian evaluation in these high dimensional problems and is also consistent with recent research (e.g. \cite{liu2017quasi}) that shows quasi-Newton methods are often preferred for physics-based simulations.

\begin{table}
\caption{Performance comparisons using the projective Newton's method. For dynamic simulations, we use a timestep of $50$~ms (i.e., 20 fps).}
\label{table-projective-newton-statistics}
\small
\begin{tabular}{c|c|c|cc}
\hline
\multirow{2}{*}{Example} & \multirow{2}{*}{Type} & \multirow{2}{*}{Sim. Type} & \multicolumn{2}{c}{Simulation step time (ms)} \\ \cline{4-5} 
                         &                       &                            & \multicolumn{1}{c|}{Ours}    & Vanilla        \\ \hline
Bistable                 & Unsup.                & Dyn.                       & \multicolumn{1}{c|}{2.2}     & 2.6 (1.2x)    \\ \hline
Cloth                    & Unsup.                & Dyn.                       & \multicolumn{1}{c|}{5.9}     & 10.7 (1.8x)   \\ \hline
Twist Bar                & Sup.                  & Static                     & \multicolumn{1}{c|}{6.3}     & 10.4 (1.7x)    \\ \hline
Dinosaur                 & Sup.                  & Dyn.                       & \multicolumn{1}{c|}{102.5}   & 340.6 (3.3x)   \\ \hline
Elephant                 & Sup.                  & Dyn.                       & \multicolumn{1}{c|}{79.8}    & 175.4 (2.2x)    \\ \hline
Bunny                    & Sup.                  & Dyn.                       & \multicolumn{1}{c|}{38.4}    & 125.3 (3.3x)   \\ \hline
\end{tabular}
\end{table}

\subsection{Discussion on Runtime Cubature Acceleration}
As mentioned in Sec.~\ref{sec-cubature-method}, while conventional methods (e.g. \cite{fulton2019latent, shen2021high, trusty2023subspace}) utilize cubature accelerations during runtime to achieve fast estimation of subspace gradients, we use cubatures solely to facilitate the training process. This is due to two reasons, the first is the negative impact of using runtime cubature accelerations on simulation accuracy (as shown in \cite{fulton2019latent}), the second is that we find the acceleration provided by runtime cubatures on the GPU backend is modest.

We demonstrate this by comparing the performance of our method with the vanilla method in the bar example, using two different backends (CPU and GPU), and simulating both with and without the use of runtime cubatures.
As data presented in Table~\ref{table-bar-cubature-comparison} show, our method results in similar acceleration rates across all four settings, indicating that the acceleration effects of our method are independent of backends and runtime cubature use. Meanwhile, the acceleration rates brought by runtime cubatures are consistent across the same backend for both our method and the vanilla method (i.e., $\sim$$1.1\times$ on GPU and $\sim$$9.4\times$ on CPU). This suggests that the acceleration effects of our method and runtime cubatures can stack. On the CPU backend, by combining our method with the cubature method, the acceleration rate can reach $25.3\times$ (28.3 ms v.s. 715.0 ms). On the other hand, the benefits of combining runtime cubature on GPU are modest ($\sim$$1.1\times$) in contrast to that in training ($\sim$$45.9\times$). This is because GPUs already provide sufficient parallelism for the full gradient evaluation when running \textit{a single} simulation, but not for the training process running in a batch. Therefore cubature acceleration is not applied during runtime but for the training process in this work.

\begin{table}[h]
\caption{Performance comparisons for runtime cubature acceleration on the bar example. The CPU results are obtained using an Intel(R) Xeon(R) Silver 4216 CPU. The number of runtime cubatures is 300.}
\label{table-bar-cubature-comparison}
\small
\begin{tabular}{c|c|cc}
\hline
\multirow{2}{*}{Backend} & \multirow{2}{*}{\begin{tabular}[c]{@{}c@{}}Runtime\\ Cubatures\end{tabular}} & \multicolumn{2}{c}{Simulation Step Time (ms)} \\ \cline{3-4} 
                         &                                    & \multicolumn{1}{c|}{Ours}    & Vanilla         \\ \hline
\multirow{2}{*}{GPU}     & No                       & \multicolumn{1}{c|}{$4.4$}     & $12.3$ ($2.78\times$)    \\ \cline{2-4} 
                         & Yes                         & \multicolumn{1}{c|}{$3.9$}     & $10.9$ ($2.76\times$)    \\ \hline
\multirow{2}{*}{CPU}     & No                       & \multicolumn{1}{c|}{$295.0$}   & $715.0$ ($2.42\times$)   \\ \cline{2-4} 
                         & Yes                         & \multicolumn{1}{c|}{$28.3$}    & $86.9$ ($3.07\times$)    \\ \hline
\end{tabular}
\end{table}

\clearpage

\end{document}